\pdfoutput=1

\documentclass[11pt]{article}

\usepackage[]{emnlp2021}
\usepackage{siunitx}
\usepackage{subcaption,booktabs}
\usepackage{dcolumn}
\newcolumntype{d}[1]{D{.}{.}{#1}}
\newcommand\mc[1]{\multicolumn{1}{c}{#1}} 
\usepackage{times}
\usepackage{latexsym}
\usepackage{multirow}
\usepackage[T1]{fontenc}

\usepackage[utf8]{inputenc}

\usepackage{microtype}
\usepackage{graphicx}

\title{mMARCO: A Multilingual Version of the MS MARCO\\Passage Ranking Dataset}

\author{Luiz Bonifacio \\
  Univ. of Campinas \\
  NeuralMind  \\ \And
  Vitor Jeronymo \\
  Univ. of Campinas \\
  NeuralMind  \\ \And
  Hugo Queiroz Abonizio \\
  NeuralMind \\ \AND
  Israel Campiotti \\
  NeuralMind  \\ \And
  Marzieh Fadaee \\
  Zeta Alpha\\ \And
  Roberto Lotufo \\
  Univ. of Campinas \\
  NeuralMind \\ \And
  Rodrigo Nogueira \\
  Univ. of Campinas\\
  Univ. of Waterloo \\
  NeuralMind \\
 }

\begin{document}

\maketitle

\begin{abstract}
The MS MARCO ranking dataset has been widely used for training deep learning models for IR tasks, achieving considerable effectiveness on diverse zero-shot scenarios. However, this type of resource is scarce in languages other than English. In this work, we present mMARCO, a multilingual version of the MS MARCO passage ranking dataset comprising 13 languages that was created using machine translation. We evaluated mMARCO by finetuning monolingual and multilingual reranking models, as well as a multilingual dense retrieval model on this dataset. We also evaluated models finetuned using the mMARCO dataset in a zero-shot scenario on Mr. TyDi dataset,  demonstrating that multilingual models finetuned on our translated dataset achieve superior effectiveness to models finetuned on the original English version alone. Our experiments also show that a distilled multilingual reranker is competitive with non-distilled models while having 5.4 times fewer parameters. Lastly, we show a positive correlation between translation quality and retrieval effectiveness, providing evidence that improvements in translation methods might lead to improvements in multilingual information retrieval. The translated datasets and finetuned models are available at \url{https://github.com/unicamp-dl/mMARCO}.
\end{abstract}

\section{Introduction}
When working with Information Retrieval (IR) tasks, one can rely on using bag-of-words (BOW) systems such as the BM25 or different approaches supported by deep learning models as dense retrieval or using a reranking pipeline. While for the for the first approach, bi-encoder models are largely used \cite{DBLP:journals/corr/abs-2004-12832, DBLP:journals/corr/abs-2007-00808,karpukhin2020dense,10.1007/978-3-030-72113-8_10,zhan2020repbert,macavaney2020expansion} to separately encode the query and the documents, in the traditional reranking pipeline an initial retrieval system (e.g., using a BOW or even a dense method) provides a list of candidates which are typically reranked using a cross-encoder model~\cite{DBLP:journals/corr/abs-2003-06713, DBLP:journals/corr/abs-1910-14424,qu2021rocketqa,zhang2021comparing,ma2021prop}. Although different, these approaches share a common aspect: the models used for both usually need to be finetuned on a labeled dataset containing queries and examples of relevant documents.


For many languages, the available training and evaluation datasets are biased towards traditional techniques \cite{DBLP:journals/corr/abs-2104-08663}, such as bag-of-words, as they are often used to build these resources~\cite{buckley2007bias,yilmaz2020reliability}. As a consequence, neural models are at a disadvantage when evaluated on these datasets. Moreover, creating new labeled resources for this task is an expensive process, resulting a lack of reliable datasets for finetuning and evaluating models for IR tasks in many languages. Some techniques to tackle the lack of labeled data include using lexical overlap or heuristics \cite{min-etal-2020-syntactic}, cross-lingual alignments \cite{DBLP:journals/corr/abs-2010-12309}, and dataset translation \cite{DBLP:journals/corr/abs-2010-11934} which is the method we adopted in this work.
As most languages have none or a limited number of labeled datasets, using automatic translation is an attainable way to derive the same resources for a wider range of languages.

More recently, researchers observed that multilingual models finetuned on monolingual datasets (usually English) achieve good effectiveness in a zero-shot manner when evaluated on other languages \cite{conneau2020unsupervised, wu2019beto,DBLP:journals/corr/abs-2010-11934, DBLP:journals/corr/abs-1912-13080}. Yet, many languages are underrepresented and the evaluation process is most of the time limited to monolingual datasets. We believe that having a high-quality labeled resource available in multiple languages allows researchers and practitioners to explore different aspects of the design space such as model architectures and training algorithms. Additionally, a way to further explore multilingual model capabilities beyond zero-shot learning is to finetune them on multilingual data. 
Considering this, we adopted an automatic translation approach to create a multilingual version of the MS MARCO passage ranking dataset, named mMARCO. This dataset is a suitable candidate as it shows good transfer learning capabilities, as well as being a popular choice for evaluating deep learning models \cite{DBLP:journals/corr/abs-2003-07820, DBLP:journals/corr/abs-2102-07662}. By translating the dataset we are able to leverages extensive manually annotated data. Lastly, as far as we know, no previous work has translated a large IR dataset such as the MS MARCO dataset to multiple languages.

Our contributions are translating the MS MARCO (both training and evaluation sets) to 13 different languages such as Portuguese, Spanish, German, and Chinese. We finetuned mono and multilingual deep learning-based models both considering dense retrieval and reranking approaches using our translated dataset. Moreover, we evaluated these models in a zero-shot scenario on the Mr. TyDi dataset, showing that multilingual models finetuned on our translated dataset achieve superior effectiveness to models finetuned on the original English version alone. Additionally, we made available the mMARCO dataset with 13 languages and all models finetuned on our translated dataset. One of them is a distilled multilingual reranker with a reduced number of parameters that is competitive with a much larger model. Others have shown that distilled models perform well on IR tasks~\cite{DBLP:journals/corr/abs-2008-09093,gao2020understanding,lu2020twinbert,hofstatter2020improving}. Our experiments confirm those results in a multilingual setup.

\section{Related Work}

With the introduction of the Transformer architecture, a variety of research was done on pretraining strategies and models. As a result of this, several multilingual pretrained models have emerged in recent years. These models extend the progress on NLP to a wide range of languages and investigate whether using a diverse set of languages during language model pretraining is beneficial when finetuning on downstream tasks. 
Models such as mBERT \cite{DBLP:journals/corr/abs-1810-04805}, XLM \cite{DBLP:journals/corr/abs-1901-07291}, XLM-R \cite{DBLP:journals/corr/abs-1911-02116}, mT5 \cite{DBLP:journals/corr/abs-2010-11934}, and mMiniLM \cite{DBLP:journals/corr/abs-2002-10957} achieved performance improvements in cross-lingual tasks, showing to be competitive with strong monolingual models. This evidence motivated \citet{DBLP:journals/corr/abs-2104-10441} to investigate whether using automatically translated datasets is sufficient in these tasks and there is no need to train monolingual models on various languages.

There are several successful attempts in the literature reporting the use of dataset translation techniques. \citet{Rodrigues2019MultilingualTE} demonstrated that translating sentences to English in order to use an English-based pretrained model outperforms its multilingual counterpart for semantic similarity and textual entailment tasks. In other studies, the dataset translation goal is not to replace multilingual models, but to provide more training data for downstream tasks that can be used in finetuning. \citet{DBLP:journals/corr/abs-1912-05200} proposed a method called Translate Align Retrieve for automatically translating the SQuAD dataset. Using the translated resource, they finetuned a multilingual BERT, and show significant improvements over previous works. In a related approach, \citet{ARAUJO20201078} demonstrated that using machine translation to translate examples to English in order to use an English monolingual model culminates in better results when compared to monolingual models in languages other than English. In addition, there are cross-lingual datasets such as XQuAD \cite{DBLP:journals/corr/abs-1910-11856}, which consists of paragraphs and question-answer pairs from the development set of SQuAD v1.1 translated into ten languages by professional translators and XNLI \cite{DBLP:journals/corr/abs-1809-05053}, an extension of NLI corpus into 15 languages. Finally, \citet{DBLP:journals/corr/abs-2010-11934} and \citet{DBLP:journals/corr/abs-2105-13626} showed that finetuning multilingual models on translated NLI and question answering datasets improved results.

In the IR community, there have been several efforts to create multilingual resources. Initiatives like TREC\footnote{\url{https://trec.nist.gov/}}, CLEF\footnote{\url{http://www.clef-initiative.eu/}}, FIRE\footnote{\url{http://fire.irsi.res.in/fire}}, and NTCIR\footnote{\url{http://research.nii.ac.jp/ntcir/}} have proposed annotated collections to evaluate IR systems. However, given the small size of most of these resources, they are not appropriate to finetune large models such as transformer-based rerankers. Moreover, the text data used to build such collections often comes from specific domains. This lack of diversity ends up harming the model's generalization ability. On the other hand, as the MS MARCO dataset assembles documents from different sources and styles, the translation process maintains this diversity in the target languages.

Closest to our work is \textsc{Mr. TyDi}~\cite{zhang2021mr}, a multilingual IR dataset derived from the TyDi QA dataset~\cite{clark2020TyDi}.
A key difference is that Mr. TyDi uses Wikipedia as a corpus, whereas mMARCO's corpus consists of diverse passages extracted from web pages.
Another difference is related to query-document relevant pairs for training per language. Mr. TyDi has an irregular distribution of examples over all languages. The Korean language has the minimum number of examples, 1,317, while the Arabic language has the maximum, 12,377. The dataset mean of training examples is 4,466. mMARCO contains 532,761 query-passage relevant pairs for each language.

\section{Methodology} 
In this section we describe the procedure to translate the MS MARCO dataset, the target languages that were selected and the two translation methods. 

\subsection{Dataset}
\noindent{\textbf{MS MARCO:}} We use the MS MARCO passage ranking dataset \cite{bajaj2016ms}, a large-scale IR dataset comprising more than half million anonymized questions that were sampled from Bing's search query logs. 
The MS MARCO dataset is formed by a collection of 8.8M passages, approximately 530k queries, and at least one relevant passage per query, which were selected by humans. The development set of MS MARCO comprises more than 100k queries. However, a smaller set of 6,980 queries is actually used for evaluation in most published works.

Regarding its format, the passages from the MS MARCO dataset do not hold any label within the text, that is, the labels are invariant to the sentence structure and the tokens ordering. This makes this dataset an excellent candidate for translation, as most labeled datasets have their labels linked to one token or a span of tokens. Since the sentence structure can vary in different languages, it is very challenging to retain the same annotation structure after translating. For instance, \citet{DBLP:journals/corr/abs-2105-06813} has demonstrated this difficulty when translating a question answering dataset. To preserve the labels on the original data, special answer delimiter symbols were added before the translation, expecting to mark the target tokens to be extracted after translation. However, the authors argue that this strategy did not work consistently and a considerable portion of the dataset was discarded after translation.
Furthermore, another reason for translating the MS MARCO dataset comes from the good transfer learning capabilities that models finetuned on MS MARCO have demonstrated \cite{DBLP:journals/corr/abs-2104-08663, Pradeep2020H2olooAT}. We believe that having the same resource available for a wider range of languages can be a unique contribution, as this dataset is arguably one of the most popular in evaluating deep learning models for \textit{ad hoc} retrieval.


\noindent{\textbf{Mr. TyDi:}} We also use Mr. TyDi dataset for benchmarking the zero-shot capabilities of our trained models. Mr. TyDi is a multilingual open retrieval extension to the TyDi dataset, with over 58M passages from Wikipedia distributed unevenly across 12 languages, 7 of which are not included in mMARCO dataset. 
Mr TyDi documents are queries were not created through translation.
Even though Mr. TyDi has a training set, we do not use it, i.e., our models are trained on mMARCO and evaluated directly on Mr. TyDi, which characterizes these as \textit{zero-shot} experiments.

\subsection{Translation}

To translate the MS MARCO dataset, we experimented with two different approaches. The first one uses the translation models made available on HuggingFace by The Language Technology Research Group at University of Helsinki~\citet{TiedemannThottingal:EAMT2020}.\footnote{https://huggingface.co/Helsinki-NLP} We refer to these models as ``Helsinki'' in the rest of this paper.
We selected the target languages according to three criteria: the largest number of Wikipedia articles, the most spoken languages in the world\footnote{https://www.ethnologue.com/guides/ethnologue200}, and the availability of the language translation pair from English to the target language in the Helsinki repository. We ended up selecting 13 languages for translation.

Although there are considerable differences between the selected languages, the translation process was conducted the same way for all of them. First, we split MS MARCO passages into sentences. As the longest passage of MS MARCO has 362 words, splitting the passages is a way to alleviate the possibility of long inputs as most translation models tend to produce high-quality translations when prompted with sentences instead of paragraphs. Once the sentences are translated to the target language, we reassemble the passage by joining all translated sentences that were previously separated. Regarding the queries, we simply translate them without any pre-processing, given that they are shorter.  
We translate batches of 64 sentences and a maximum sequence length of 512 tokens on a Tesla V100 GPU. 

The second translation approach uses Google Translate, available via a paid API~\footnote{\url{https://cloud.google.com/translate}}. Unlike the Helsinki translation, there is no need to split the passages into smaller sentences. The only requirement is the maximum number of characters, which is set to 5000. This way, we translate the passages (and queries) by grouping them in batches up to this limit of characters.
The average translation time for both translation methods is reported in Table~\ref{tab:translation_time}.

\begin{table}[h]
\centering
\begin{tabular}{l l  *{2}{d{3.3}}}
    \cline{2-4}
    & & \mc{\textbf{Passages}} & \mc{\textbf{Queries}} \\
    \cline{2-4}
    (1) & Helsinki          &  79.58  & 8.19  \\
    (2) & Google Translate  &  78.59  & 1.01  \\
    \cline{2-4}
\end{tabular}
\caption{Average translation time in hours for translating all 8.8M passages and 530K queries of the MS MARCO dataset.} 
\label{tab:translation_time}
\end{table}

After translating the dataset to all target languages, we created a multilingual training set split using an equal proportion of samples from each language. The original MS MARCO training set consists of 39 million triples. Each triple is formed by a query followed by a relevant and non-relevant passage. These triples are used to finetune a reranking model. To finetune a multilingual model using the same triples from the original dataset but distributed among different languages, we construct a set of multilingual triples. To do this, we select the same queries and passages in the original dataset and randomly replace them with their translations. We always use the same language in the triple (i.e., the query and its respective positive and negative passages). Although we translated MS MARCO into 13 different languages, the multilingual training set was created using 9 languages, including English. This allows us to evaluate the models in a zero-shot manner, i.e., finetune in one language and evaluate on another.

\begin{table*}[ht]
    \centering
    \begin{tabular}{l l c c | c c c c}
    \toprule
    & &  \multicolumn{2}{c|}{\textbf{R@1k}} & \multicolumn{4}{c}{\textbf{MRR@10}} \\
    & \textbf{Language}  & \textbf{BM25} & \textbf{mColB.} & \textbf{BM25} & \textbf{mT5} & \textbf{mMiniLM} & \textbf{mColB.} \\ 
    \midrule
    (1)  & English (Orig.) & 0.857 & 0.953 & 0.184 & 0.366 & 0.366  & 0.352 \\
    (2)  & Spanish         & 0.770 & 0.897 & 0.158 & 0.314 & 0.309  & 0.301 \\
    (3)  & French          & 0.769 & 0.891 & 0.155 & 0.302 & 0.296  & 0.289 \\
    (4)  & Italian         & 0.753 & 0.888 & 0.153 & 0.303 & 0.291  & 0.292 \\
    (5)  & Portuguese      & 0.744 & 0.887 & 0.152 & 0.302 & 0.289  & 0.292 \\
    (6)  & Indonesian      & 0.767 & 0.854 & 0.149 & 0.298 & 0.293  & 0.275 \\
    (7)  & German          & 0.674 & 0.867 & 0.136 & 0.289 & 0.278  & 0.281 \\
    (8)  & Russian         & 0.685 & 0.836 & 0.124 & 0.263 & 0.251  & 0.250 \\
    (9)  & Chinese         & 0.678 & 0.837 & 0.116 & 0.249 & 0.249  & 0.246 \\
    \midrule
    \multicolumn{6}{l}{\textit{Zero-shot (models were fine-tuned on the 9 languages above)}}\\
    (10) & Japanese        & 0.714 & 0.806 & 0.141 & 0.267 & 0.263  & 0.236 \\
    (11) & Dutch           & 0.694 & 0.862 & 0.140 & 0.292 & 0.276  & 0.273 \\
    (12) & Vietnamese      & 0.714 & 0.719 & 0.136 & 0.256 & 0.247  & 0.180 \\
    (13) & Hindi           & 0.711 & 0.785 & 0.134 & 0.266 & 0.262  & 0.232 \\
    (14) & Arabic          & 0.638 & 0.749 & 0.111 & 0.235 & 0.219  & 0.209 \\
    \bottomrule
    \end{tabular}
    \caption{Main results on the mMARCO passage ranking dataset. The rerankers \textsc{mT5} and \textsc{mMiniLM} and the dense retrieval model mColBERT were fine-tuned on mMARCO translated with Google Translate.}
    \label{tab:baselines_bm25}
\end{table*}

\begin{table*}[ht]
    \centering\centering\resizebox{1.0\textwidth}{!}{
    \begin{tabular}{rlrrrrrrrrrrr}
    & Translation & \textit{es} & \textit{fr} & \textit{pt} & \textit{it} & \textit{id} & \textit{de} & \textit{ru} & \textit{zh} & \textit{ar} & \textit{hi} & \textbf{avg} \\
    \toprule
    \multicolumn{2}{l}{\textbf{BM25}} \\
    (1) & Helsinki & 0.144 & 0.138 & 0.141 & 0.131 & 0.120 & 0.121 & 0.083 & 0.064 & 0.089 & 0.016 & 0.105 \\
    (2) & Google & 0.158 & 0.155 & 0.152 & 0.153 & 0.149 & 0.136 & 0.124 & 0.116 & 0.111 & 0.134 & 0.138\\
    \midrule
    \multicolumn{2}{l}{\textbf{mMiniLM}} \\
    (3) & Helsinki & 0.292 & 0.271 & 0.275 & 0.252 & 0.239 & 0.258 & 0.174 & 0.142 & 0.169 & 0.035 & 0.211 \\
    (4)   & Google & 0.309 & 0.296 & 0.289 & 0.291 & 0.293 & 0.278 & 0.251 & 0.249 & 0.219    & 0.262    & 0.274 \\
    \midrule
    \multicolumn{2}{l}{\textbf{mT5}} \\
    (5) & Helsinki & 0.297 & 0.279 & 0.285 & 0.248 & 0.244 & 0.264 & 0.183 & 0.152 & 0.187 & 0.035 & 0.217 \\
    (6) & Google & 0.314 & 0.302 & 0.302 & 0.303 & 0.298 & 0.289 & 0.263 & 0.249 & 0.235 & 0.266 & 0.281 \\
    \bottomrule
    \end{tabular}
    }
    \caption{Comparison of Helsinki translation models (open source) vs Google Translate (commercial). The reported metric is MRR@10 on the development set of mMARCO.}
    \label{tab:helsinki_vs_google}
\end{table*}

\subsection{Experimental Setup}
We evaluate the datasets derived from our translation process in the passage ranking task. Following a two-stage pipeline widely adopted in IR \cite{DBLP:journals/corr/abs-1910-14424}, we first retrieve a ranked list of translated passages using BM25 \cite{DBLP:journals/corr/abs-2101-05667, DBLP:journals/corr/abs-2003-06713, DBLP:journals/corr/abs-2008-09093} with the translated queries as input. We use the Pyserini framework to perform this task~\cite{lin2021pyserini}.

We further rerank the list of passages using multilingual pretrained \textsc{mT5} and \textsc{mMiniLM} models. These models were finetuned on our multilingual training set, comprising examples in 9 different languages. To finetune the \textsc{mT5} model, we follow the recommendation of ~\citet{DBLP:journals/corr/abs-2003-06713}. While the aforementioned work used \texttt{true} and \texttt{false} as predictions tokens, we train our \textsc{mT5} model to generate a \texttt{yes} token when a given document is relevant to a query and a \texttt{no} token otherwise. We use a batch size of 128, a maximum sequence length of 512 tokens, and finetune the model for 100k iterations, which took approximately 27 hours on a TPU v3-8. We use a constant learning rate of 0.001 and all layers use a dropout of 0.1.

The \textsc{mMiniLM} model was finetuned with a batch size of 32 and a maximum sequence length of 512 tokens, which took approximately 50 hours on a Tesla A100 GPU. A learning rate warm-up was set over the 5,000 first training batches. We use the implementation provided by~\citet{reimers-2020-Curse_Dense_Retrieval}.
For both finetuning procedures, we use an equal number of examples per language. The \textsc{mT5} was trained with 12.8 million training examples, while \textsc{mMiniLM} used 80 million examples. 
For inference, both models rerank the top 1000 passages retrieved by BM25.

Furthermore, we finetune and evaluate a multilingual dense model based on ColBERT~\cite{khattab2020colbert}, referred to as mColBERT in this paper. We start from the mBERT checkpoint, which was pretrained on approximately one hundred languages. We then finetune it on mMARCO's translation from Google Translate with a batch size of 64, and maximum sequence length of 180 and 32 tokens for passages and queries respectively. This model is finetuned with 25.6M examples, which takes approximately 36 hours on a Tesla V100 GPU. We set mColBERT's final linear layer dimension to 64, but we split each token embedding into two, thus resulting in an embedding of size 32 and twice the amount of embeddings for retrieval. Compared to the original ColBERT's implementation that uses embeddings of size 128, our method achieves a similar MRR@10 on the MS MARCO development set while using half of the CPU memory. Cosine similarity was used in both training and retrieval.

Regarding the evaluation on Mr. TyDi's test set, its documents are already segmented into passages. However, some passages still exceed the maximum input sequence length of mColBERT and mT5. Hence, for both models, these passages were further segmented into windows of 10 sentences and a stride of 5. For that, we used Spacy's tokenizer in their respective languages. The only exception where the passages were used as they were is Swahili, as there was no Swahili sentence tokenizer available on Spacy. Furthermore, the base model for our mColBERT finetuning (bert-multilingual-uncased) was not pretrained on Thai. Hence, Thai results are not shown.

The English corpus is the largest in Mr. TyDi, with almost 33M passages. Indexing with mColBERT requires about a day on a single NVIDIA V100 and occupies 442GB of space. Note that this amount is loaded into CPU's memory for fast retrieval. 
BM25 was indexed with the original passages as it is not limited by a maximum sequence length.  When retrieving 1000 documents per query, we were able to successfully reproduce the results obtained by~\citet{zhang2021mr}. mT5 took on average 3 hours per language to perform reranking on a NVIDIA V100.

\section{Results} \label{section:results}
The main results on the development set of mMARCO are shown in Table~\ref{tab:baselines_bm25}. We report MRR@10, which is the official metric of the MS MARCO passage dataset, as well as recall@1000. As our baseline, row (1) shows results from the original dataset in English, i.e., no translation involved.

\begin{table*}[ht]
    \centering\resizebox{1.0\textwidth}{!}{
\begin{tabular}{lcrrrrrrrrrrr} 
\hline
\multicolumn{1}{l}{} & \multicolumn{1}{c}{finetuning}& \textit{ar} & \textit{bn} & \textit{en}  & \textit{fi} & \textit{id} & \textit{ja} & \textit{ko} & \textit{ru} & \textit{sw} & \textit{te} & \textbf{avg}  \\
\hline

\multicolumn{13}{c}{\textbf{MRR@100}}\\

BM25  & - & 0.368 & 0.418 & 0.140 & 0.284 & 0.376 & 0.211 & 0.285 & 0.313 & 0.389 & 0.343 & 0.313 \\
mT5 & EN & 0.625 & 0.621 & 0.341 & 0.479 & 0.609 & 0.454 & 0.460 & 0.520 & 0.623 & 0.671 & 0.532 \\
mT5 & MULTI & 0.622 & 0.651 & 0.357 & 0.495 & 0.611 & 0.481 & 0.474 & 0.526 & 0.629 & 0.666 & 0.551 \\
mColBERT & MULTI & 0.553 & 0.488 & 0.329 & 0.413 & 0.555 & 0.366 & 0.367 & 0.482 & 0.448 & 0.616 & 0.461 \\
\hline
\multicolumn{13}{c}{\textbf{Recall@100}}\\

BM25 & - & 0.793 & 0.869 & 0.537 & 0.719 & 0.843 & 0.645 & 0.619 & 0.648 & 0.764 & 0.758 & 0.720 \\
mT5 & EN & 0.893 & 0.936 & 0.719 & 0.853 & 0.927 & 0.825 & 0.779 & 0.764 & 0.841 & 0.852 & 0.829 \\
mT5 & MULTI & 0.884 & 0.923 & 0.724 & 0.851 & 0.928 & 0.832 & 0.765 & 0.763 & 0.838 & 0.850 & 0.835 \\
mColBERT & MULTI & 0.859 & 0.918 & 0.786 & 0.826 & 0.911 & 0.709 & 0.729 & 0.861 & 0.808 & 0.969 & 0.837 \\
\hline
\end{tabular}
    }
\caption{Main results on the Mr. TyDi passage ranking dataset. \textsc{mT5} and  \textsc{mColBERT} were finetuned on mMARCO translated with Google Translate.}
\label{tab:mr_TyDi}
\end{table*}

\begin{table*}[ht]
    \centering
    \begin{tabular}{c c  c  c  c  c c}
    \cline{2-7}
    & \textbf{pretraining} & \textbf{finetuning} & \multicolumn{ 2}{c}{\textbf{EN}} & \multicolumn{ 2}{c}{\textbf{PT}} \\
    \cline{2-7}
    & & & \textbf{T5} & \textbf{MiniLM} & \textbf{T5} &\textbf{MiniLM} \\
    \cline{4-7}
    (1) & EN & EN    & 0.381 & 0.396 & 0.181 &   0.164   \\
    (2) & PT & PT    & 0.200 &   -   & 0.299 &   -       \\
    (3) & PT & EN+PT & 0.354 &   -   & 0.301  &   -      \\
    \cline{2-7}
    (4) &\multirow{ 4}{*}{MULTI}  & EN      & 0.371 & 0.382  & 0.293 &   0.277   \\
    (5) &  & PT      & 0.357 &   0.336   & 0.303 & 0.296 \\
    (6) &  & EN+PT   & 0.374 & 0.374 & 0.306 & 0.299 \\
    (7) &  & MULTI   & 0.366 & 0.366 & 0.302 & 0.277 \\
    \cline{2-7}
    \end{tabular}
    \caption{Ablation results for monolingual and multilingual models. The reported metric is MRR@10.}
    \label{tab:results_mono_multi}
\end{table*}

The first observation is regarding R@1000 of BM25, in which the English dataset has the highest among all, while an average drop of 0.14 is observed in other languages. One possible reason for this drop is the lexical mismatch between the translated query and relevant passages.
Since queries and passages are translated independently, the same word can be translated into two different (but synonym) words in the translated query and passage.
For instance, the English word \texttt{car} may be translated as \texttt{carro} or \texttt{automóvel} when translated to Portuguese. If because of the context, the former translation is used in the relevant passage and the latter is used in the translated query, a lexical mismatch is introduced in the resulting dataset. As a consequence, BM25 might not return this relevant document as it relies on lexical matching.
Subsequently, rerankers might be penalized. 

On the other hand, mColBERT's R@1000  suggests that a dense retrieval model mitigates the lexical mismatch problem. For all languages, it achieved higher figures when compared to BM25. For instance, German (7) and Dutch (11) mColBERT's R@1000 are 0.193 and 0.168 points higher than BM25's.

Considering the results from English evaluation, we observe that \textsc{mT5} and \textsc{mMiniLM} achieve the same MRR@10. This is an interesting finding since \textsc{mMiniLM} is a much lighter model; while \textsc{mT5} has 580 million parameters, \textsc{mMiniLM} has 107 million. This result further shows the effectiveness of the \textsc{mMiniLM} language model distillation. 
Comparing the remaining results between \textsc{mT5} and \textsc{mMiniLM}, we observe an average drop of 0.08 and 0.09 in MRR@10 for rows (2) to (14) when compared to the English results. 

The results from rows (10) to (14) came from a zero-shot evaluation, i.e., they were finetuned on the nine languages shown in rows (1) to (9) and directly evaluated on the languages in rows (10) to (14). These results show how effective are multilingual models, both considering language and task aspects.

Table~\ref{tab:helsinki_vs_google} compares the IR results when using open source and commercial translation mechanisms. We observe that all languages have benefited from Google Translate translations when considering the MRR@10 metric. Moreover, we highlight the improvements on languages like Russian, Chinese, Arabic, and Hindi, as the commercial translations resulted in improvements in MRR@10 scores.

\subsection{Zero-shot results}
Results on the Mr. TyDi dataset are shown on Table~\ref{tab:mr_TyDi}. The official metrics for this dataset are MRR@100 and R@100.

In all languages, models trained on mMARCO outperform BM25, which was the strongest single model reported by \citet{zhang2021mr}.
More importantly, the mT5 reranker finetuned on mMARCO (marked as ``MULTI'' in the ``Finetuning'' column) outperforms, on average, mT5 finetuned only on the original English MS MARCO. 
This is the case even for languages that were not present in mMARCO, such as Bengali, Finnish, Japanese, Korean and Swahili.
This means that our translated multilingual training dataset is beneficial in a zero-shot setting as well.
Similar to mMARCO's development set results, mT5 outperforms mColBERT in MRR in all languages, and interestingly, English has the smallest difference of them all.

\subsection{Ablation Study}
In this section, we verify how monolingual and multilingual models perform when varying pretraining, finetuning, and evaluation languages. As the number of pairwise language combinations is large, we only use English and Portuguese in this ablation study. The monolingual models we use are an English~\cite{DBLP:journals/corr/abs-1910-10683} and a Portuguese~\cite{carmo2020ptt5} pretrained T5 model.

Table~\ref{tab:results_mono_multi} shows the results. Rows (1) to (3) use models mostly pretrained on a single language (T5 and miniLM), and rows (4) to (7) report results for multilingual pretrained models (\textsc{mT5} and \textsc{mMiniLM}). When finetuned on monolingual datasets, we observe insignificant differences between monolingual and multilingual models in both English and Portuguese evaluation sets. We observe that for the monolingual configuration, \textsc{miniLM} outperforms \textsc{T5} in English (1). 
We argue that the model was able to leverage training data for both languages and thus reduced most inaccuracies introduced by noisy translations. 

When considering the results from English dataset, it is important to mention that both multilingual rerankers were finetuned on a smaller amount of English examples than the models in row (1).
This shows that the translation did not harm the resulting datasets significantly. Albeit the lexical mismatch problem discussed in Section~\ref{section:results}, both rerankers were able to learn the task from the translated data.

Rows (4) and (5) in Table~\ref{tab:results_mono_multi} exhibit the results for \textsc{mT5} finetuned on monolingual datasets. 
Whilst the results were lower for the English dataset when compared to monolingual \textsc{T5} (1), the results on the Portuguese version were marginally higher than the ones observed in row (2).
This cross-lingual evaluation shows how a multilingual model finetuned on one language and evaluated on another still can reach good results. Although this observation is not true for the Portuguese-English cross-lingual evaluation (row 5), as the result was below the monolingual one (row 1), the difference is small. Even when finetuned on a translated dataset, the multilingual model was able to achieve competitive results when evaluated on the original English dataset. Additionally, multilingual T5 finetuned only on Portuguese (row 5) outperformed Portuguese models (rows 2 and 3) when evaluated on Portuguese.
This result supports the observation that multilingual models finetuned on monolingual datasets perform considerably well on the same task in a different language \cite{conneau2020unsupervised, wu2019beto,DBLP:journals/corr/abs-2010-11934}. 

The best result on the Portuguese MS MARCO was achieved by finetuning \textsc{mT5} and \textsc{mMiniLM} on the English-Portuguese versions of MS MARCO. On the other hand, the same models achieve lower results when finetuned on more than two languages, as shown in row (7). The results are even lower for \textsc{mMiniLM}. Given the small number of parameters of this model, we hypothesize that it is not able to fully benefit from other languages.

Lastly, the models finetuned on multilingual data were not far below when evaluated on English and Portuguese datasets. Particularly for \textsc{mT5}, the multilingual finetuned model (row 7) slightly outperformed the monolingual one (row 2) in the Portuguese dataset. Once more, this evidence supports our hypothesis that our translated dataset can be beneficial for multilingual models during finetuning.

\subsection{Translation Quality vs Retrieval Effectiveness }
\begin{figure}[h]
  \centering
  \includegraphics[width=8cm]{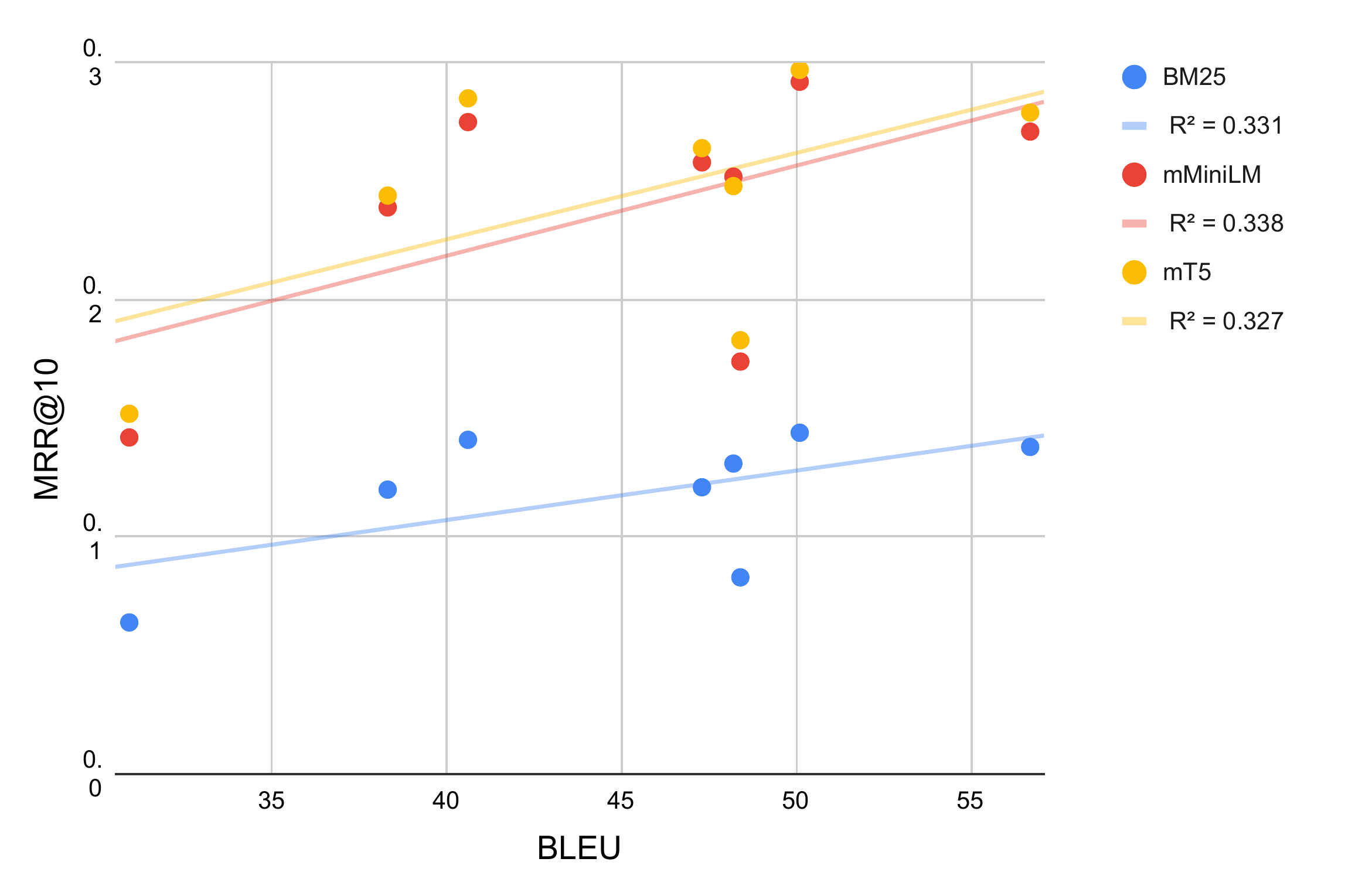}
  \caption{Translation quality measured as BLEU on Tatoeba vs retrieval quality measured as MRR@10 on mMARCO.}
  \label{fig:bleu_mrr}
\end{figure}

In this section, we investigate the correlation between the quality of the translation models, measured in BLEU points, and the effectiveness of different retrieval models on mMARCO, measured in MRR@10.
In Figure~\ref{fig:bleu_mrr}, the x-axis represents the BLEU scores of Helsinki translation models on the Tatoeba dataset~\cite{artetxe2019massively}. The source language is English and the \texttt{target} language is one of \{es, fr, it, pt, id, de, ru, zh\} languages. The y-axis is the MRR@10 of the retrieval methods (BM25, \textsc{mMiniLM} and \textsc{mT5}) on the mMARCO subset of that same language.
The three trend lines have a $R^2$ of approximately 0.33, which shows a weak correlation between translation quality and retrieval effectiveness.
Thus, it is reasonable to expect that improvements in translation methods can bring improvements to multilingual information retrieval.

\section{Conclusions and Future Work}
In this work, we translate and make available mMARCO, a multilingual IR dataset in 13 different languages. This resource can be used for training and evaluating models.
Additionally, we train and evaluate several monolingual and multilingual Transformer-based models on these datasets and provide benchmarks for further study in multilingual IR. 
As a way to encourage future creation of more datasets in new languages, we made available our translation code. In addition, all models and datasets are available at HuggingFace. 

Our findings indicate that multilingual models finetuned on multilingual datasets achieve competitive results when compared to monolingual approaches (both in pretraining and finetuning approaches). Furthermore, we show that the translation quality has a significant impact on retrieval results. This outcome showed to be even greater when considering non-Latin based languages, such as Russian and Chinese, where the commercial translation achieved higher results in contrast to translations from open sourced models. Our results also showed that translating datasets is a feasible mechanism to overcome the labeled data scarcity.

Furthermore, we demonstrated how a lighter distilled model, \textsc{miniLM}, is competitive when finetuned in the same way as a much larger model. As future work, we would like to evaluate our finetuned models on a dataset in a language never seen during the language model pretraining or finetuning. 


\bibliographystyle{ACM-Reference-Format}
\bibliography{ref}
\end{document}